%% file: lrec-coling2024-example.tex
\documentclass[10pt, a4paper]{article}
\usepackage{lrec-coling2024} 

\usepackage{listings}
\usepackage{microtype}
\usepackage{booktabs}
\usepackage{amsmath}
\usepackage{bm}
\usepackage{enumitem}
\usepackage{graphicx}
\usepackage{svg}
\usepackage{tablefootnote}
\usepackage{amsfonts}
\usepackage[symbol]{footmisc}
\usepackage{multirow}
\usepackage{hyperref}
\usepackage{cleveref}
\usepackage{algorithm}
\usepackage{algpseudocode}
\usepackage{amssymb}
\usepackage{algorithmicx,eqparbox,array}
\usepackage{natbib}
\usepackage{multibib}
\usepackage{subcaption}
\usepackage{float}
\usepackage{perpage}

\makeatletter
\def\@mb@citenamelist{cite,citep,citet,citealp,citealt,citepalias,citetalias}
\makeatother
\newcites{languageresource}{~}

\usepackage{graphicx}
\usepackage{tabularx}
\usepackage{soul}
\usepackage{titlesec}
\newcommand{\RNum}[1]{\lowercase\expandafter{\romannumeral #1\relax}}

\usepackage{xcolor}
\usepackage{tikz}
\usetikzlibrary{calc}
\usepackage{hyperref}
\definecolor{darkblue}{rgb}{0, 0, 0.5}
\hypersetup{colorlinks=true, citecolor=darkblue, linkcolor=darkblue, urlcolor=darkblue}

\usepackage{xstring}
\usepackage{pifont}
\usepackage{bold-extra}
\usepackage{anyfontsize}

\usepackage{stfloats}
\usepackage{arydshln}
\usepackage{hhline}

\let\oldReturn\Return
\renewcommand{\Return}{\State\oldReturn}

\definecolor{color0}{HTML}{e38e03} 
\definecolor{color1}{HTML}{5f97f7} 
\definecolor{color2}{HTML}{cf6759} 
\definecolor{color3}{HTML}{99ae38} 
\definecolor{deepgreen}{rgb}{0.0, 0.5, 0.0} 

\title{\makebox[1pt][r]{\includegraphics[height=1.4em]{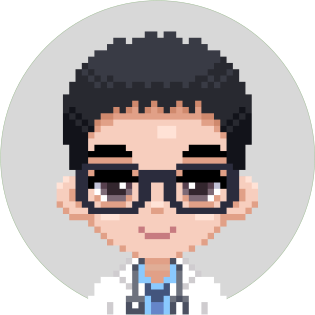}}\,Dr3: Ask Large Language Models Not to Give Off-Topic Answers in Open Domain Multi-Hop Question Answering}

\name{Yuan Gao$^{\spadesuit, \clubsuit}$, Yiheng Zhu$^\clubsuit$, Yuanbin Cao$^\clubsuit$, Yinzhi Zhou$^\clubsuit$, Zhen Wu$^{\spadesuit\dagger}$\thanks{$^\dagger$Corresponding Author}, \\ \textbf{\large Yujie Chen$^\clubsuit$, Shenglan Wu$^\clubsuit$, Haoyuan Hu$^\clubsuit$, Xinyu Dai$^\spadesuit$}} 
\address{$^\spadesuit$National Key Laboratory for Novel Software Technology, Nanjing University\\
         $^\clubsuit$Artificial Intelligence Department, Cainiao Network \\
         gaoy@smail.nju.edu.cn, \{wuz, daixinyu\}@nju.edu.cn\\
         \{zyh171911, lingzun.cyb, yinzhi.zyz, aisling.cyj, shenglan.wsl,  haoyuan.huhy\}@cainiao.com}
\abstract{
Open Domain Multi-Hop Question Answering (ODMHQA) plays a crucial role in Natural Language Processing (NLP) by aiming to answer complex questions through multi-step reasoning over retrieved information from external knowledge sources.
Recently, Large Language Models (LLMs) have demonstrated remarkable performance in solving ODMHQA owing to their capabilities including planning, reasoning, and utilizing tools.
However, LLMs may generate off-topic answers when attempting to solve ODMHQA, namely the generated answers are irrelevant to the original questions.
This issue of off-topic answers accounts for approximately one-third of incorrect answers, yet remains underexplored despite its significance.
To alleviate this issue, we propose the Discriminate$\rightarrow$Re-Compose$\rightarrow$Re-Solve$\rightarrow$Re-Decompose (Dr3) mechanism.
Specifically, the Discriminator leverages the intrinsic capabilities of LLMs to judge whether the generated answers are off-topic.
In cases where an off-topic answer is detected, the Corrector performs step-wise revisions along the reversed reasoning chain (Re-Compose$\rightarrow$Re-Solve$\rightarrow$Re-Decompose) until the final answer becomes on-topic.
Experimental results on the HotpotQA and 2WikiMultiHopQA datasets demonstrate that our Dr3 mechanism considerably reduces the occurrence of off-topic answers in ODMHQA by nearly 13\%, improving the performance in Exact Match (EM) by nearly 3\% compared to the baseline method without the Dr3 mechanism$^1$\thanks{$^1$ Our code and data will be available at \url{https://github.com/Gy915/Dr3}.}.
\newline 
}

\begin{document}

\maketitleabstract

\begin{table*}[h!]
\setlength\heavyrulewidth{0.16em}
\centering
\resizebox{\linewidth}{!}{
\large
\begin{tabular}{llllcc}
\toprule
\textbf{No.} & \bf{Question}& \bf{Gold Answer} & \bf{Pred Answer} & \bf{Correct} & \bf{On-Topic} \\
\midrule
\multirow{2}{*}{1} & \multirow{2}{*}{Who is older Danny Green or James Worthy?} & \multirow{2}{*}{James Worthy} & \multirow{2}{*}{James Worthy} & \multirow{2}{*}{{\color{color3}\ding{52}}} & \multirow{2}{*}{{\color{color3}\ding{52}}} \\
& & & & & \\
\multirow{2}{*}{2} & \multirow{2}{*}{``Tunak'' is a pop love song by an artist born in which year?} & \multirow{2}{*}{1997} & \multirow{2}{*}{1967} & \multirow{2}{*}{{\color{color2}\ding{56}}} & \multirow{2}{*}{{\color{color3}\ding{52}}} \\
& & & & & \\
\multirow{2}{*}{3} & Which film was released more recently, The Secret Life Of & \multirow{2}{*}{The Secret Life Of Pets 2} & \multirow{2}{*}{Lover 3}  & \multirow{2}{*}{{\color{color2}\ding{56}}} & \multirow{2}{*}{{\color{color2}\ding{56}}} \\
& Pets 2 or Love Me Deadly? & & & & \\
\multirow{2}{*}{4} & \multirow{2}{*}{What languages did the son of Sacagawea speak?} & \multirow{2}{*}{French and English} & \multirow{2}{*}{Edinburgh} & \multirow{2}{*}{{\color{color2}\ding{56}}} & \multirow{2}{*}{{\color{color2}\ding{56}}} \\
& & & & & \\
\bottomrule
\end{tabular}
}
\caption{Examples of off-topic answers. In the 1st example, the predicted answer is both correct and on-topic. In the 2nd example, the answer is incorrect but still on-topic. In the 3rd and 4th examples, the answers are not only incorrect but also off-topic.}
\label{tab:example}
\vspace*{-0.5em}
\end{table*}

\input{0_introduction}
\input{1_preliminary}

\input{2_method}

\input{3_experiment}
\input{4_result_and_analysis}

\input{5_related_work}
\input{6_conclusion}

\nocite{*}
\section{Bibliographical References}
\label{sec:reference}

\bibliographystyle{lrec_natbib}
\bibliography{lrec-coling2024-example}


\appendix

\newpage

\input{7_appendix}

\end{document}

%% file: 0_introduction.tex
\section{Introduction}

Open Domain Multi-Hop Question Answering (ODMHQA) is one of the most challenging tasks in Natural Language Processing (NLP) \citep{mavi2022survey}.
Unlike Reading Comprehension (RC) tasks that provide paired contexts \citep{rajpurkar2016squad}, ODMHQA operates in an open-domain setting thereby requiring models to retrieve contexts from external knowledge sources like Wikipedia \citep{feldman2019multi}.
Unlike single-hop QA where answers can be derived from a single source, ODMHQA exhibits greater complexity as final answers require reasoning over multiple sources in a multi-hop fashion \citep{zhu2021retrieving}.
Therefore, ODMHQA is more realistic than basic QA as it involves multi-hop reasoning over the retrieved contexts in an open-domain setting.

Nowadays, Large Language Models (LLMs) have become a \textit{de facto} choice for solving ODMHQA \citep{wei2022chain,press2022measuring,wang2023plan}.
Among recent works, \citet{yao2022react} proposed the ReAct paradigm in which LLMs are prompted to solve complex problems, inspiring a series of subsequent studies \citep{hao2023toolkengpt,hsieh2023tool,ruan2023tptu}.
ReAct prompts LLMs to generate both reasoning traces and actions to interact with the external world in an interleaved manner, outperforming vanilla acting models \citep{nakano2021webgpt} while being competitive with pure reasoning approaches \citep{wei2022chain}.

However, LLMs encounter the issue of generating \textbf{off-topic answers} when solving ODMHQA.
Specifically, an off-topic answer refers to when the generated answer is irrelevant to the original question.
For example, the answer ``Barack Obama'' is an off-topic answer to the question ``In which year was David Beckham's wife born?'', where an on-topic answer should be a year rather than a name (more examples can be found in Table \ref{tab:example}).
The process of solving ODMHQA intrinsically involves steps of problem reasoning, task planning, and tool utilization.
During these steps, the generation and accumulation of irrelevant information due to inherent hallucinations of LLMs \citep{zhang2023hallucination,bang2023multitask} can result in off-topic answers.
In fact, the analysis presented in Section \ref{ssec:prevalence_of_off-topic_answers} reveals that approximately 1/3 of the incorrect answers are identified as off-topic answers, but this issue has not been sufficiently explored so far.

In this paper, to reduce the occurrence of off-topic answers, we propose the Discriminate$\rightarrow$ Re-Compose$\rightarrow$Re-Solve$\rightarrow$Re-Decompose (Dr3) mechanism that performs post-hoc judgment and subsequently corrects the reasoning chain through backtracking.
Specifically, the Discriminator leverages the intrinsic capabilities of LLMs to determine whether the generated answer is off-topic relative to the original question.
In cases where an off-topic answer is detected, the Corrector backtracks and revises the reasoning chain (Re-Compose$\rightarrow$Re-Solve$\rightarrow$Re-Decompose).

Experimental results on the HotpotQA and 2WikiMultiHopQA datasets demonstrate that our Dr3 mechanism considerably improves the performance of LLMs on ODMHQA.
Additionally, we conduct dedicated studies to \RNum{1}) investigate the capability of the Discriminator in capturing off-topic answers and its impact on the consequent correctness of ODMHQA, \RNum{2}) examine the impacts of three individual components of the Corrector (Re-Compose, Re-Solve, and Re-Decompose) on reducing off-topic answers, \RNum{3}) explore the effect of sub-question numbers on off-topic answers, and \RNum{4}) investigate the issue of off-topic answers across different types of questions.

\vspace*{0.5em}

To sum up, our contribution is threefold:
\vspace*{-0.2em}
\begin{enumerate}[leftmargin=0em, itemindent=1.5em]
    \setlength\itemsep{-0.2em}
    \item To the best of our knowledge, we are the first to point out and analyze the issue of off-topic answers in solving ODMHQA using LLMs.
    \item We propose the Dr3 mechanism to reduce the occurrence of off-topic answers in solving ODMHQA using LLMs, which contains a Discriminator to detect off-topic answers by LLMs and a Corrector to heuristically correct the off-topic answers.
    \item We conduct extensive experiments on the HotpotQA and 2WikiMultiHopQA datasets to demonstrate the effectiveness of our Dr3 mechanism. The results show that Dr3 reduces the occurrence of off-topic answers by nearly 13\%, while improving the question answering performance by nearly 3\% in Exact Match (EM) compared to ReAct.
\end{enumerate}

%% file: 1_preliminary.tex
\section{Preliminary}
\label{sec:preliminary}
In this section, we begin by introducing the ReAct approach, which serves as the basic framework for solving ODMHQA using LLMs.
Because the reasoning and planning steps are intertwined within the thought component of vanilla ReAct, which hinders identifying and fixing potential issues for ODMHQA, we modify ReAct to establish an equivalent variant called ReAct+ (marginally outperforming the unmodified version), where we explicitly decouple the sub-questions within the reasoning chain for ODMHQA.
Subsequently, we elaborate on the issue of off-topic answers that arise when solving ODMHQA using ReAct+.
Furthermore, we provide statistics on the error types associated with off-topic answers, offering insights that inspire our proposed method in the next section.

\subsection{ReAct(+)}
\label{ssec:react}

\textbf{Vanilla ReAct.} To solve ODMHQA using LLMs, \citet{yao2022react} proposed ReAct that uses interleaved steps of reasoning and acting.
Specifically, ReAct verbally maintains high-level plans for acting (reason to act), while interacting with external environments such as Wikipedia to incorporate additional information into reasoning (act to reason).
Formally, ReAct interacts with the environment in $N$ steps before concluding the final answer.
At step $i$, based on the previous step's \textbf{observation}, $o_{i-1} \in \mathcal{O}$ (where $\mathcal{O}$ is the set of all possible observations from the environment), ReAct develops a \textbf{thought}, $\tau_i \in \mathcal{T}$ (where $\mathcal{T}$ is the set of all possible LLM-generated thoughts), by reasoning over the available information, and planning the next step hence deliver an \textbf{action}, $a_i \in \mathcal{A}$ (where $\mathcal{A}$ is the set of optional actions like \texttt{\textbf{search}}[\texttt{query}] to retrieve passages or \texttt{\textbf{finish}}[\texttt{answer}] to conclude the question), leading to a new observation, $o_{i} \in \mathcal{O}$, from the environment.
Specifically, the action is selected by ReAct's \textbf{policy}, $\pi: H_i \rightarrow a_i$, where the \textbf{context} $H_i \equiv (o_0, \tau_1, a_1, o_1, \ldots, \tau_{i-1}, a_{i-1}, o_{i-1}, \tau_i)$ represents the interaction history between ReAct and the environment.


\begin{figure}[thb]
\includegraphics[width=\linewidth]{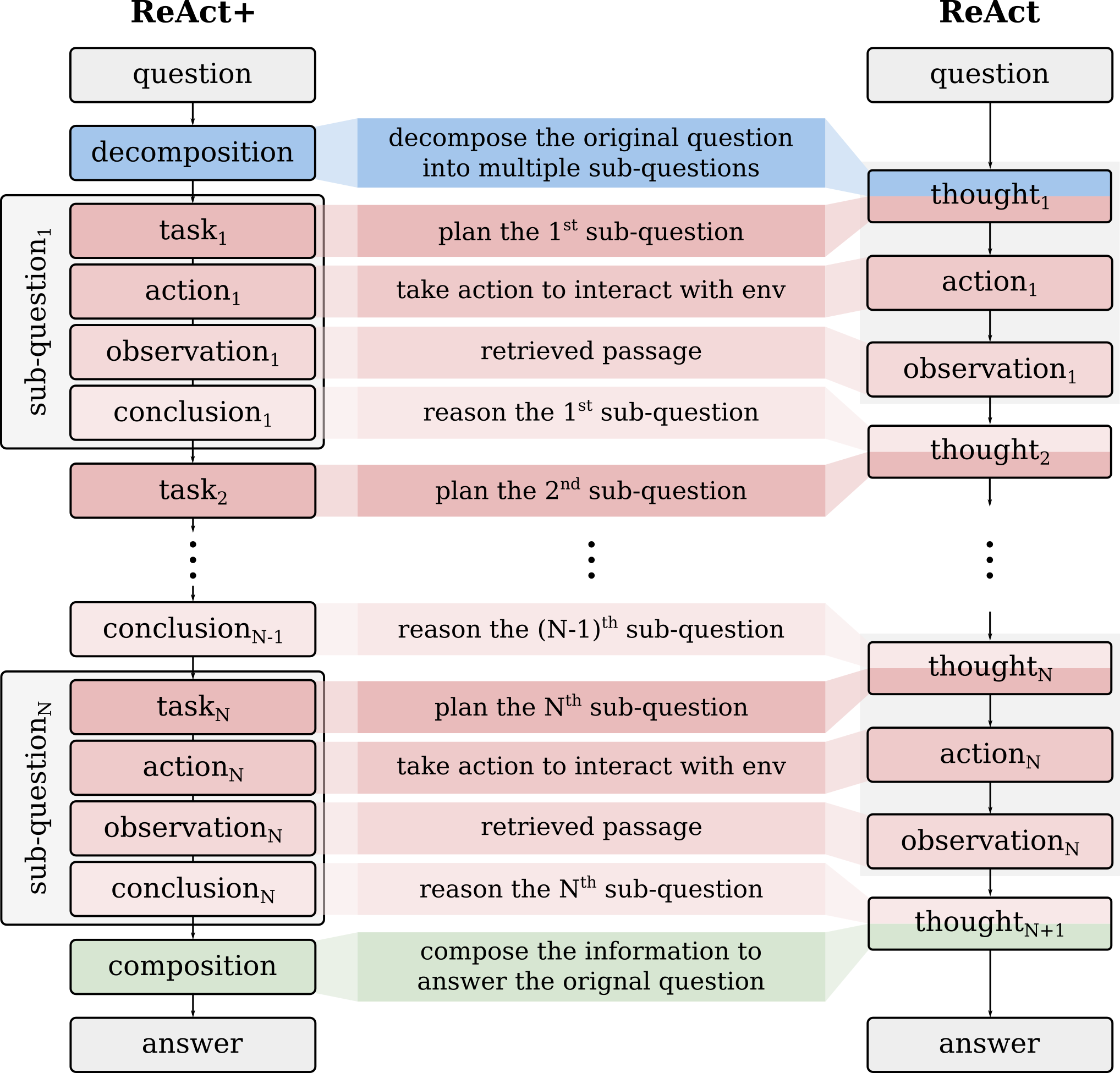}
\caption{ReAct+: An equivalent variant of ReAct for open domain multi-hop question answering, where the sub-questions are explicitly decoupled.}
\label{fig:react_plus}
\vspace*{-0.5em}
\end{figure}

\textbf{Modified ReAct.} 
It is noteworthy that ReAct's reasoning and planning steps are intertwined within its thought component, denoted as $\tau_i$ at step $i$.
This entanglement presents challenges in identifying potential issues when solving ODMHQA, which hinders further optimization.
Accordingly, we develop ReAct+ (shown in Figure \ref{fig:react_plus}), an equivalent variant of ReAct better suited for multi-hop reasoning chains in ODMHQA.
ReAct+ decomposes a complex question into a set of clear-cut \texttt{Sub-Questions}, inspired by \citet{press2022measuring}.
In ReAct+, the \texttt{Sub-Question} at step $i$ is denoted by $S_i \equiv (t_i, a_i, o_i, c_i)$, where $t_i$ signifies the planned \textbf{task}, $a_i$ represents the selected \textbf{action}, $o_i$ represents the \textbf{observation} from the environment, and $c_i$ symbolizes the intermediate \textbf{conclusion}.
As Figure \ref{fig:react_plus} illustrates, ReAct's $\tau_i$ corresponds to ReAct+'s $(c_{i-1}, t_i)$ for intermediate \texttt{Sub-Questions}.
Regarding the first and last \texttt{Sub-Questions}, $\tau_1$ corresponds to $(D, t_1)$ while $\tau_{N+1}$ corresponds to $(c_N, C)$, where $D$ denotes \textbf{\texttt{Decomposition}} and $C$ represents \textbf{\texttt{Composition}}.
By implementing these modifications, we observe that ReAct+ marginally outperforms ReAct on ODMHQA (see Section \ref{ssec:dr3_results}), indicating that decomposing a complex question into \texttt{Sub-Questions} does not impair performance, aligning with the findings of \citet{mishra2022reframing}.

\subsection{Prevalence of Off-Topic Answers}
\label{ssec:prevalence_of_off-topic_answers}




\begin{figure}[htb]
\vspace*{-1em}
\includegraphics[width=\linewidth]{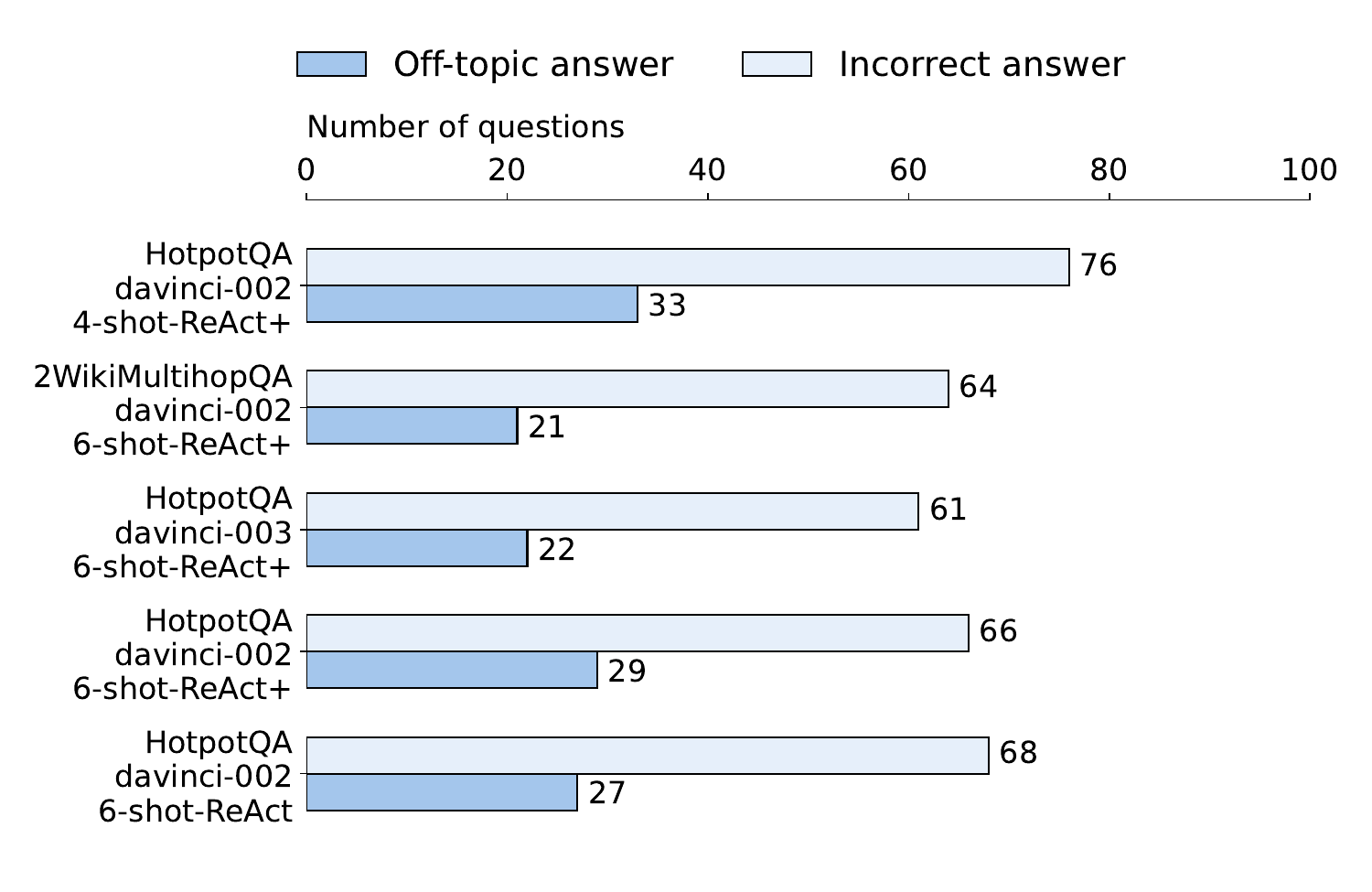}
\vspace*{-2em}
\caption{Statistics on off-topic and incorrect answers in ODMHQA with LLMs. We evaluated 100 randomly sampled cases for each combination of different datasets, LLMs, prompts, and methods.}
\label{fig:mismatch}
\end{figure}

\begin{figure*}[pt]
\hspace*{-2.4em}
\begin{subfigure}{.5\textwidth}
  \centering
  \includegraphics[width=.8\linewidth]{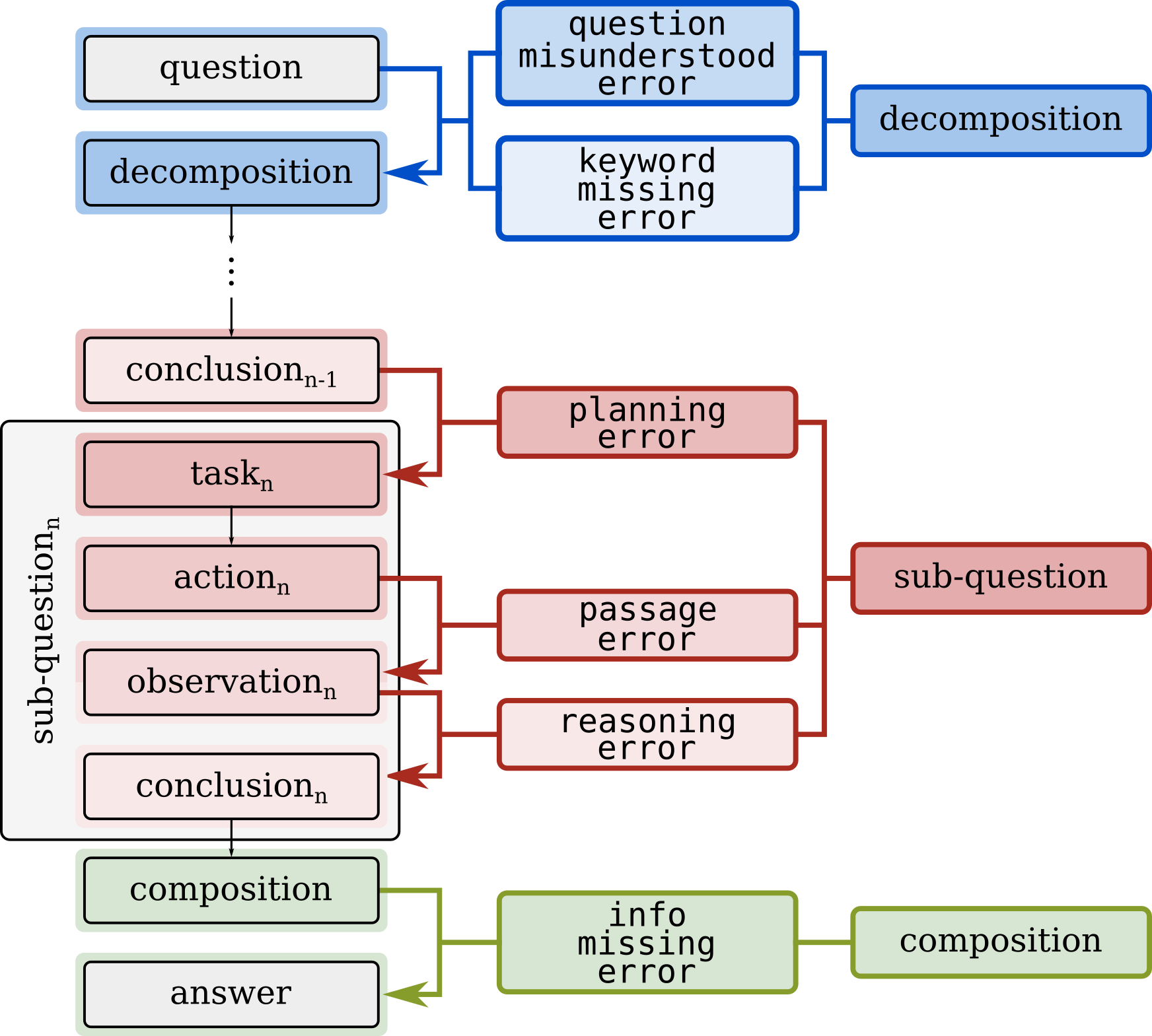}
\end{subfigure}%
\begin{subfigure}{0.55\textwidth}
  \centering
  \includegraphics[width=1.0\linewidth,trim={0 6em 1em 5em},clip]{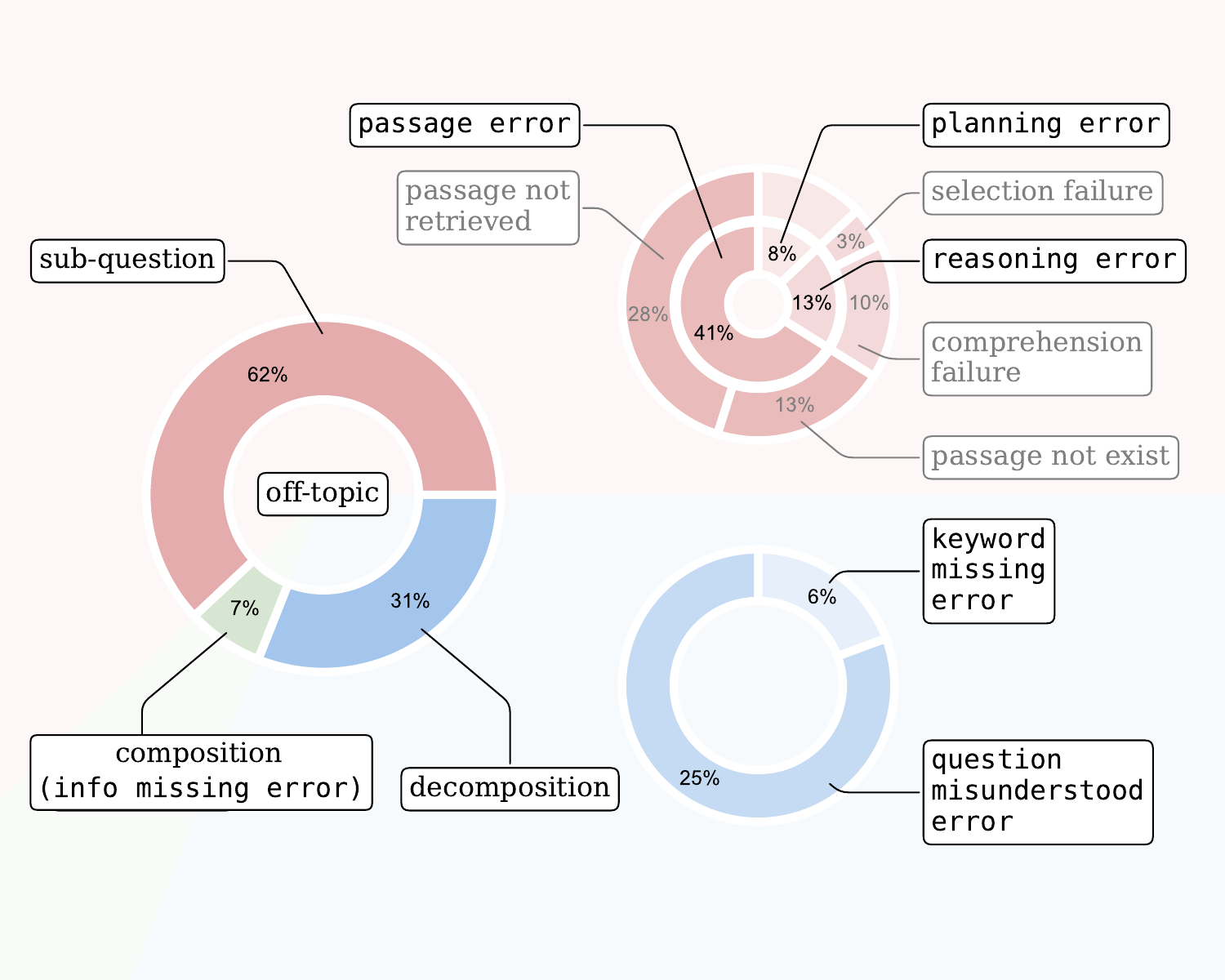}
\end{subfigure}
\caption{Cause analysis on off-topic answers. We locate and classify the causes behind 112 off-topic answers out of 500 cases randomly sampled from the HotpotQA dataset using ReAct+.}
\label{fig:error_types}
\vspace*{-0.5em}
\end{figure*}

We observe that LLMs encounter the problem of generating \textbf{off-topic answers} when attempting to solve ODMHQA.
To elaborate, off-topic answers refer to cases where the generated answers are not relevant to the original questions, which is similar to the concepts of off-topic responses in dialogue systems \citep{malinin2016off} and off-topic essays in high-stakes tests \citep{louis2010off}.
Notably, off-topic answers are necessarily incorrect answers.
We show two cases of off-topic answers in the 3rd and 4th examples in Table \ref{tab:example}.
In the 3rd example, on-topic answers should be either ``The Secret Life of Pets 2'' or ``Love Me Deadly'', as indicated in the original question.
However, the generated answer ``Lover 3'' does not match either of these expected on-topic answers.
In the 4th example, an on-topic answer should specify the language(s), while the generated answer ``Edinburgh'' is a city name instead of specifying a language.

To determine the significance of the off-topic issue, we randomly sample 500 cases from the HotpotQA \citep{yang2018hotpotqa} and 2WikiMultiHopQA \citep{ho2020constructing} datasets and manually label whether the LLM-generated answers are off-topic.
As shown in Figure \ref{fig:mismatch}, off-topic answers are a prevalent issue observed across different datasets, LLMs, prompts, and methods.
We notice that \textbf{approximately 1/3 of the incorrect answers are identified as off-topic answers}, which directly impairs the user experience of employing LLMs for solving ODMHQA.


\subsection{Cause Analysis on Off-Topic Answers}
\label{ssec:cause_analysis_on_off-topic_answers}

To further investigate the issue of off-topic answers, we analyze 112 off-topic answers out of 500 cases randomly sampled from the HotpotQA dataset \citep{yang2018hotpotqa} using ReAct+.
We thoroughly review the full solving history and then locate and classify the causes behind these off-topic answers.
For these off-topic cases, the distribution of causes identified through this analysis is shown in Figure \ref{fig:error_types}.
Here are the key findings from our analysis:
\begin{itemize}[leftmargin=0em, itemindent=1em]
    \setlength\itemsep{-0.2em}
    \item The \texttt{Decomposition} step accounts for 31\% of the off-topic answers, where LLMs misunderstand the original question or lose the keyword during the process of decomposing the question.
    \item The \texttt{Sub-Question} steps contribute to the largest portion at 62\% of the off-topic answers. We further classify these as planning errors, passage errors, and reasoning errors. Notably, passage errors \citep{yao2022react} cover 41\% of the cases, occurring when the relevant passage was not successfully retrieved or does not exist in the external database. Reasoning errors stem from two main sources: comprehension failures, where the model fails to generate the correct intermediate answer by faithfully following the reasoning chain, known as unfaithful reasoning \citep{lyu2023faithful}; and selection failures, where the model mistakenly chooses another answer candidate instead of the required one.
    \item The \texttt{Composition} step accounts for a relatively small portion at 7\% of the off-topic answers. In these cases, we observe that the LLM becomes trapped in hallucination \citep{zhang2023hallucination}, missing the key information from the original question or overlooking key evidence from intermediate conclusions, ultimately causing off-topic answers.
\end{itemize}

%% file: 2_method.tex
\section{Method}

\begin{figure*}[thb]
\centering
\includegraphics[width=\linewidth]{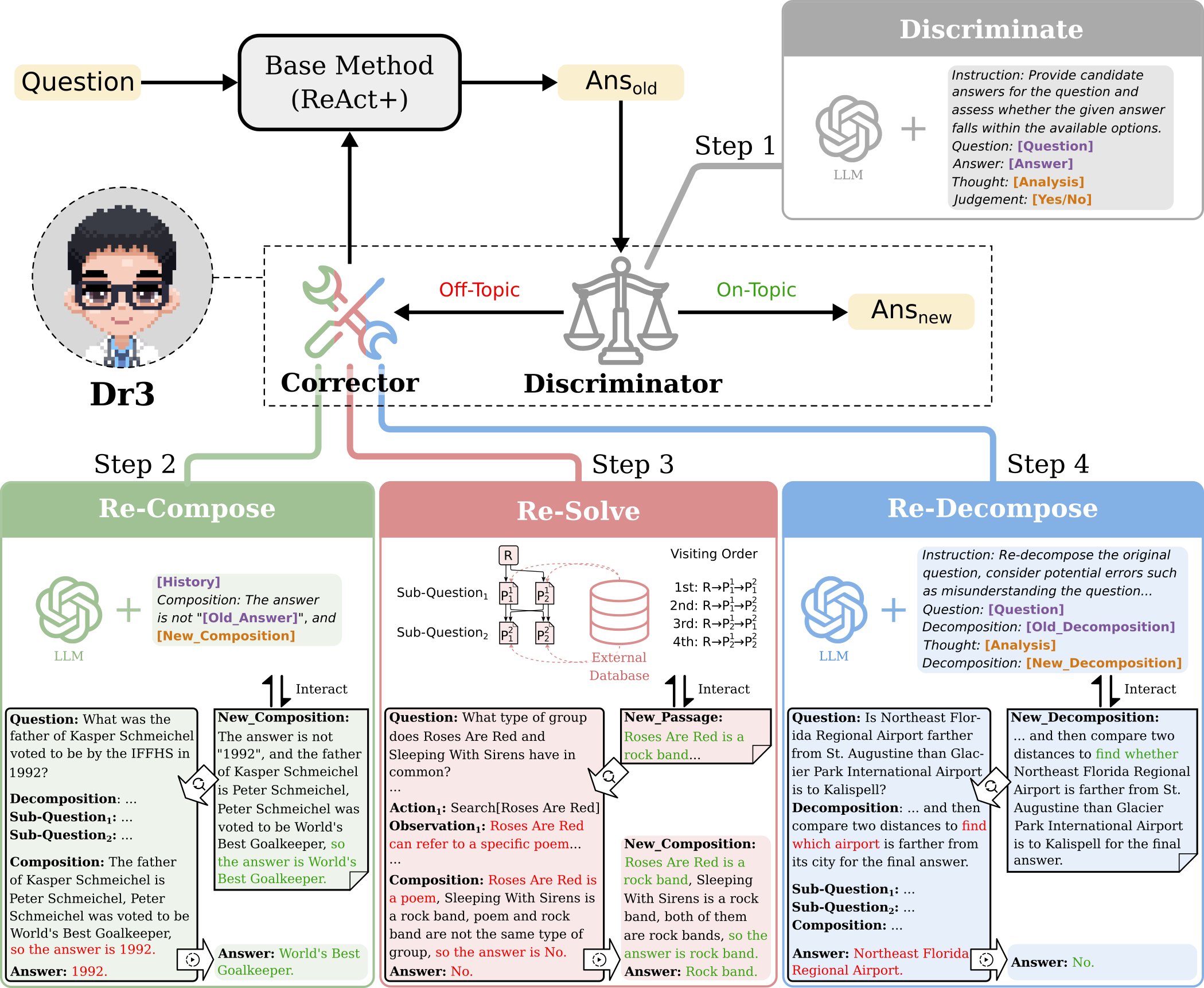}
\caption{Dr3 mechanism: Discriminate$\rightarrow$Re-Compose$\rightarrow$Re-Solve$\rightarrow$Re-Decompose.}
\label{fig:dr3}
\vspace*{-0.5em}
\end{figure*}
In this section, we present our Dr3 mechanism (demonstrated in Figure \ref{fig:dr3}), which aims to address the issue of off-topic answers, based on the analysis presented in Section \ref{ssec:cause_analysis_on_off-topic_answers}. 
The Dr3 mechanism consists of two key modules: the Discriminator and the Corrector.
The Discriminator judges whether the generated answer is on-topic by leveraging the intrinsic capabilities of LLMs.
The Corrector applies heuristic rules to correct the solving history along the reversed reasoning chain (Re-Compose$\rightarrow$Re-Solve$\rightarrow$Re-Decompose) until the Discriminator confirms that the answer is on-topic.

\subsection{Discriminator}
In the Discriminator, we leverage the intrinsic capabilities of LLMs to determine whether the generated answer is on-topic.
This is accomplished by feeding both the original question and the LLM-generated answer into the Discriminator, which yields a binary judgment of either \texttt{YES} or \texttt{NO}, indicating whether the generated answer is on-topic or off-topic.
Specifically, we design an instruction that prompts the LLM to first conceptualize candidate answers and subsequently assess whether the generated answer falls within the range of available options.
If the generated answer is among the candidate answers, it should be considered on-topic; otherwise, it should be considered off-topic.

\subsection{Corrector}

When the generated answer, $\texttt{Ans}_{\texttt{old}}$, is identified as off-topic by the Discriminator, the Corrector revises the reasoning chain in three stages.
These three stages follow the order, Re-Compose$\rightarrow$Re-Solve$\rightarrow$Re-Decompose, which mirrors ReAct+'s original order of reasoning, \texttt{Composition}$\leftarrow$\texttt{Sub-Question}$\leftarrow$\texttt{Decomposi-} \texttt{tion}.
At each revision stage, the Corrector generates a new answer, $\texttt{Ans}_{\texttt{new}}$, which is evaluated by the Discriminator for on-topic or off-topic.
This iterative process continues either until the Discriminator approves $\texttt{Ans}_{\texttt{new}}$ as an on-topic answer, or until the entire reasoning chain has been exhausted.
We introduce the Corrector's three revision stages as follows.

\vspace*{-0.5em}

\subsubsection{Re-Compose}
The Corrector's Re-Compose stage addresses the off-topic issue that occurs in ReAct+'s original \texttt{Composition} stage.
Inspired by \citet{zheng2023progressive}, we provide a hint, $h_{C}$, to the LLM, stating ``\texttt{The answer is not} $[\texttt{Ans}_{\texttt{old}}]$'' at the start of the \texttt{Composition} stage.
This hint encourages the LLM to reconsider the question and evidence, generating a new answer that is potentially on-topic.

\vspace*{-0.5em}

\subsubsection{Re-Solve}
The Corrector's Re-solve stage addresses the off-topic issue that occurs during ReAct+'s original \texttt{Sub-Question} stage.
The focus here is on fixing passage errors, which was identified as the primary cause for off-topic answers in the analysis conducted in Section \ref{ssec:cause_analysis_on_off-topic_answers}.
For \texttt{Sub-Question} $S_i$, we first replace the passage in the observation, $o_i$, based on retrieval probabilities from the IR system.
Subsequently, we utilize ReAct+ to resolve $S_i$ and obtain a new answer, $\texttt{Ans}_{\texttt{new}}$.
This procedure repeats for $S_i$ until the Discriminator approves that $\texttt{Ans}_{\texttt{new}}$ is an on-topic answer.
If the number of replacements reaches the predefined threshold, $T_D$, without generating an on-topic answer, we iterate this procedure backwards for \texttt{Sub-Question}, $S_{i-1}$.
This continues backwards until an on-topic answer is obtained or the first \texttt{Sub-Question}, $S_1$, is addressed.

\vspace*{-0.5em}

\subsubsection{Re-Decompose}
The Corrector's Re-Decompose stage addresses the off-topic issue that occurs in ReAct+'s original \texttt{Decomposition} stage.
Inspired by \citet{xi2023self}, we leverage the intrinsic capabilities of LLMs to refine the \texttt{Decomposition}.
To achieve this, we input the original question along with the existing \texttt{Decomposition} into the LLM.
Our prompts guide the LLM to revise the \texttt{Decomposition} in cases of errors including misunderstanding the question and missing keywords.
Subsequently, the LLM generates a new \texttt{Decomposition} that replaces the old one, enabling ReAct+ to continue execution and generate a new answer, $\texttt{Ans}_{\texttt{new}}$.

%% file: 3_experiment.tex
\section{Experiments}

\subsection{Datasets and Metrics}

We present the results obtained from two popular datasets: HotpotQA and 2WikiMultiHopQA.
\textbf{HotpotQA} \citep{yang2018hotpotqa} is a 2-hop QA dataset built from Wikipedia passages where reasoning chains are formed between passage pairs.
We use the same data following \citet{yao2022react} as our test dataset, which randomly samples 500 cases from Dev split.
For each case, we only provide the question without the paired passages according to the open domain setting.
\textbf{2WikiMultiHopQA} \citep{ho2020constructing} constructs multi-hop QA cases by combining Wikipedia articles and Wikidata knowledge.
Compared to HotpotQA, 2WikiMultiHopQA is more challenging because it includes four question types: comparison, inference, compositional, and bridge-comparison.
Similar to \citet{yao2022react} and \citet{khattab2022demonstrate}, we randomly sample 500 question-answer pairs from the Dev split as the testing dataset.
Following previous work in QA \citep{yang2018hotpotqa,xu2023search}, we evaluate using the metrics of token F1 score, exact match (EM), and cover exact match (Cover EM).

\subsection{Baselines}
\label{ssec:baselines}

The baseline methods used for comparison are briefly described as follows.

\begin{itemize}[leftmargin=0em, itemindent=1em]
    \setlength\itemsep{-0.2em}
    \item \textbf{Zero-Shot} \citep{kojima2022large}: The zero-shot approach requires the LLM to output the answer to the question without any relevant example.
    \item \textbf{Few-Shot} \citep{brown2020language}: The Few-Shot approach prompts the LLM with a few relevant question–answer examples, demonstrating the procedure for solving this type of task.
    \item \textbf{CoT} \citep{wei2022chain}: The Chain-of-Thought (CoT) approach facilitates the LLM to generate coherent intermediate reasoning steps that mimic a step-by-step thought process by prompting ``\texttt{Let's think step-by-step}''.
    \item \textbf{ReAct} \citep{yao2022react}: The Reasoning-Acting (ReAct) approach enhances reasoning-only LLMs by incorporating acting capabilities through the use of external tools, which defines the reasoning pattern as a sequence of interleaved Thought-Action-Observation steps (see Section \ref{ssec:react}).
    \item \textbf{ReAct+} (Ours): The ReAct+ approach modifies the reasoning pattern of ReAct to fit multi-hop question answering, which involves interleaved Task-Action-Observation-Conclusion steps (see Section \ref{ssec:react}).
\end{itemize}



\subsection{Implementation Details}
For the Re-Solve step, we set the maximum number of replaced passages $T_D$ to 3 for each \texttt{Sub-Question}.
We use the off-the-shelf ColBERTv2 \citep{santhanam2022colbertv2} following \citet{khattab2022demonstrate} as the IR system, which encodes the Wikipedia corpus by passage.
For the Re-Decompose step, we use 6 question-answer examples in prompts.
Within each step, we set the maximum number of continuous \texttt{Sub-Question} to 7, following \citet{yao2022react}.
All methods are implemented using \textit{text-davinci-002} as the LLM, in line with \citet{yao2022react} in September 2023. The detailed prompts can be found in Appendix \ref{sec:react_plus_prompt}, \ref{sec:discriminator_prompt}, and \ref{sec:re-decompose_prompt}.

%% file: 4_result_and_analysis.tex
\section{Results and Analysis}

In this section, we first present the overall results of our Dr3 mechanism.
Following that, we evaluate the performance of the Discriminator in detecting off-topic answers.
Next, we assess the Corrector on lowering the off-topic ratio.
Additionally, we analyze the relationship between the off-topic issue and factors including the number of \texttt{Sub-Questions}, and multi-hop question types.

\subsection{Main Results}
\label{ssec:dr3_results}

\begin{table}[htb]
\vspace*{-1em}
\renewcommand{\arraystretch}{1.8}
\setlength\heavyrulewidth{0.28em}
\fontsize{20pt}{20pt}\selectfont
\centering
\resizebox{\linewidth}{!}{
\begin{tabular}{l|ccc|ccc}
\toprule
&  \multicolumn{3}{c|}{\textbf{HotpotQA}} & \multicolumn{3}{c}{\textbf{2WikiMultiHopQA}} \\
\textbf{Method} & \textbf{EM}$\uparrow$ & \textbf{Cover EM}$\uparrow$ & \textbf{F1}$\uparrow$ & \textbf{EM}$\uparrow$ & \textbf{Cover EM}$\uparrow$ & \textbf{F1}$\uparrow$ \\
\midrule
Zero-Shot & 8.80 & 27.80 & 22.65 & 4.60 & 29.8 & 17.08 \\
Few-Shot & 21.20 & 28.00 & 34.41 & 20.40 & 22.80 & 25.45 \\
CoT & 30.40 & 37.60 & 43.93 & 28.80 & 33.60 & 36.37 \\
ReAct & 30.80$^\dagger$ & - & - & 34.40 & 41.80 & 43.24 \\
ReAct+ (Ours) & 31.00 & 36.60 & 42.21 & 35.60 & 43.60 & 45.39 \\
\textbf{Dr3 (Ours)} & \textbf{33.80} & \textbf{40.00} & \textbf{46.53} & \textbf{38.80} & \textbf{46.60} & \textbf{48.36} \\
\bottomrule
\end{tabular}
}
\caption{Main results on the HotpotQA and 2WikiMultiHopQA datasets. The result with $^\dagger$ is borrowed from \citet{yao2022react}. }
\label{result1} 
\end{table}

In this subsection, we evaluate the overall performance of our Dr3 mechanism on the HotpotQA and 2WikiMultiHopQA datasets, compared to five baseline methods described in Section \ref{ssec:baselines}.
As shown in Table \ref{result1}, our Dr3 approach outperforms existing methods across evaluation metrics of EM, Cover EM, and F1 score.
Specifically, compared to the best baseline ReAct+, our Dr3 approach exhibits significant improvements:
on HotpotQA, we observe absolute enhancements of 2.80\% in EM, 3.40\% in Cover EM, and 4.32\% in F1 score.
Meanwhile, on 2WikiMultiHopQA, we achieve even larger gains of 3.20\% in EM, 3.00\% in Cover EM, and 2.97\% in F1 score.
These compelling results clearly demonstrate the effectiveness of our Dr3 mechanism for improve the performance on ODMHQA.

\subsection{Performance of Discriminator}


\begin{table}[htb]
\vspace*{-0.5em}
\setlength\heavyrulewidth{0.05em}
\renewcommand{\arraystretch}{0.4}
\fontsize{3pt}{3pt}\selectfont
\centering
\resizebox{\linewidth}{!}{
\begin{tabular}{cccc}
\toprule
\textbf{Accuracy} & \textbf{Precision} & \textbf{Recall} & \textbf{F1} \\
\midrule[0.01em]
92.77 & 83.76 & 85.21 & 84.82 \\
\bottomrule
\end{tabular}
}
\caption{Performance of Discriminator in detecting off-topic answers. Human evaluation as the ground truth on the HotpotQA dataset, while precision, recall, F1 are for the off-topic class.}
\label{table:mismatch_judger}
\end{table}

\begin{figure}[htb]
\centering
\includegraphics[width=0.6\linewidth]{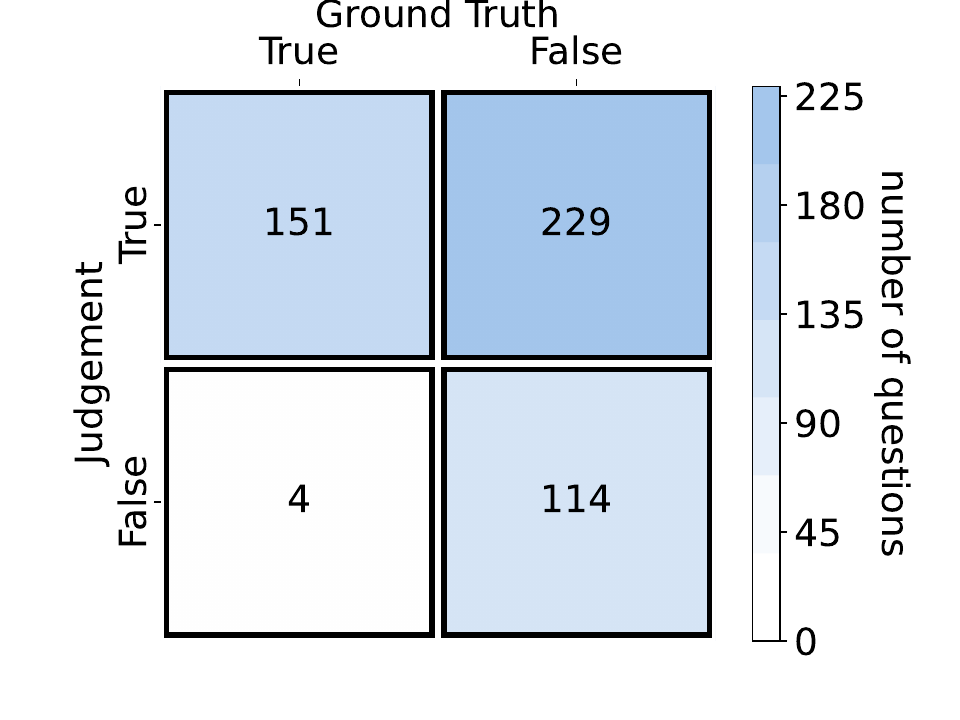}
\vspace*{-1em}
\caption{Confusion matrix of QA correctness by Discriminator. Evaluation performed on the HotpotQA dataset, while ``True'' means correct and ``False'' means incorrect.}
\label{fig:confusion_matrix}
\vspace*{-0.5em}
\end{figure}
In this subsection, we examine the performance of the Discriminator in detecting off-topic answers.
\textbf{Discriminator is comparable to human judgment in detecting off-topic answers.}
As shown in Table \ref{table:mismatch_judger}, the Discriminator achieves an accuracy of 92.77\% on the HotpotQA dataset, which manifests its remarkable capability in identifying off-topic answers, closely resembling human judgment.
Moreover, the high F1 score of 84.82\% further emphasizes the Discriminator's capability to recall a majority of the off-topic answers while maintaining a high precision.

%
%
\textbf{Discriminator identifies 1/3 off-topic answers and rarely misjudges correct answers.}
Figure \ref{fig:confusion_matrix} illustrates the confusion matrix of the Discriminator in judging QA correctness on the HotpotQA dataset.
The Discriminator has a sensitivity of 97.4\%($\frac{151}{151+4}$), indicating that the Discriminator rarely misjudges correct answers as incorrect.
Meanwhile, it has a specificity of 33.2\%($\frac{114}{114+229}$), signifying that it successfully identifies a substantial portion of incorrect answers.
Therefore, the benefit of identifying incorrect answers outweighs the misjudgments.

\subsection{Performance of Corrector}

\begin{table*}[htbp]
\renewcommand{\arraystretch}{1.8}
\setlength\heavyrulewidth{0.1em}
\fontsize{6pt}{6pt}\selectfont
\centering
\resizebox{\linewidth}{!}{
\begin{tabular}{ccc|cccc}
\toprule
\textbf{Re-Decompose} & \textbf{Re-Solve} & \textbf{Re-Compose} & \textbf{EM}$\uparrow$ & \textbf{F1}$\uparrow$ & \textbf{Off-Topic Ratio (Discriminator)}$\downarrow$ & \textbf{Off-Topic Ratio (Expert)}$\downarrow$ \\
\midrule
{\color{color2}\ding{56}} & {\color{color2}\ding{56}} & {\color{color2}\ding{56}} & 31.00 & 42.21 & 23.69 & 23.09 \\
{\color{color2}\ding{56}} & {\color{color2}\ding{56}} & {\color{color3}\ding{52}} & 32.00 & 43.11 & 18.07 & - \\
{\color{color2}\ding{56}} & {\color{color3}\ding{52}} & {\color{color3}\ding{52}} & 33.40 & 45.51 & 9.83 & - \\
{\color{color3}\ding{52}} & {\color{color3}\ding{52}} & {\color{color3}\ding{52}} & \textbf{33.80} & \textbf{46.53} & \textbf{7.42} & \textbf{10.24} \\
\bottomrule
\end{tabular}
}
\caption{Ablation studies of the Corrector modules. Evaluation performed on the HotpotQA dataset.}
\label{table:corrector_ablation_study} 
\vspace*{-0.5em}
\end{table*}

In this subsection, we conduct ablation studies on three individual Corrector modules to investigate their effectiveness in improving question answering and alleviating the off-topic issue.
The results are shown in Table \ref{table:corrector_ablation_study}.
\textbf{Re-Solve is the most effective module, exhibiting a 1.4\% EM improvement.}
This significant EM improvement can be attributed to Re-Solve identifying passage errors as the primary cause of off-topic answers (see Section \ref{ssec:cause_analysis_on_off-topic_answers}).
%
\textbf{Re-Compose (+1.0\% EM) appears to be more effective than Re-Decompose (+0.4\% EM).}
This is unexpected that Re-Compose is so effective given \texttt{Composition} errors comprise the smallest ratio (7\%) as shown in Section \ref{ssec:cause_analysis_on_off-topic_answers}.
We conjecture this occurs because Re-Compose not only fixes \texttt{Composition} errors but also encourages the LLM to re-examine the full solving history.
By thoroughly reassessing the question and evidence, Re-Compose can implicitly fix reading comprehension and \texttt{Decomposition} errors as well. To demonstrate this phenomenon, we present 2 cases in Appendix \ref{sec:special_case}.

\subsection{The Effect of Sub-Question Numbers}

\begin{figure}[htb]
\centering
\includegraphics[width=\linewidth]{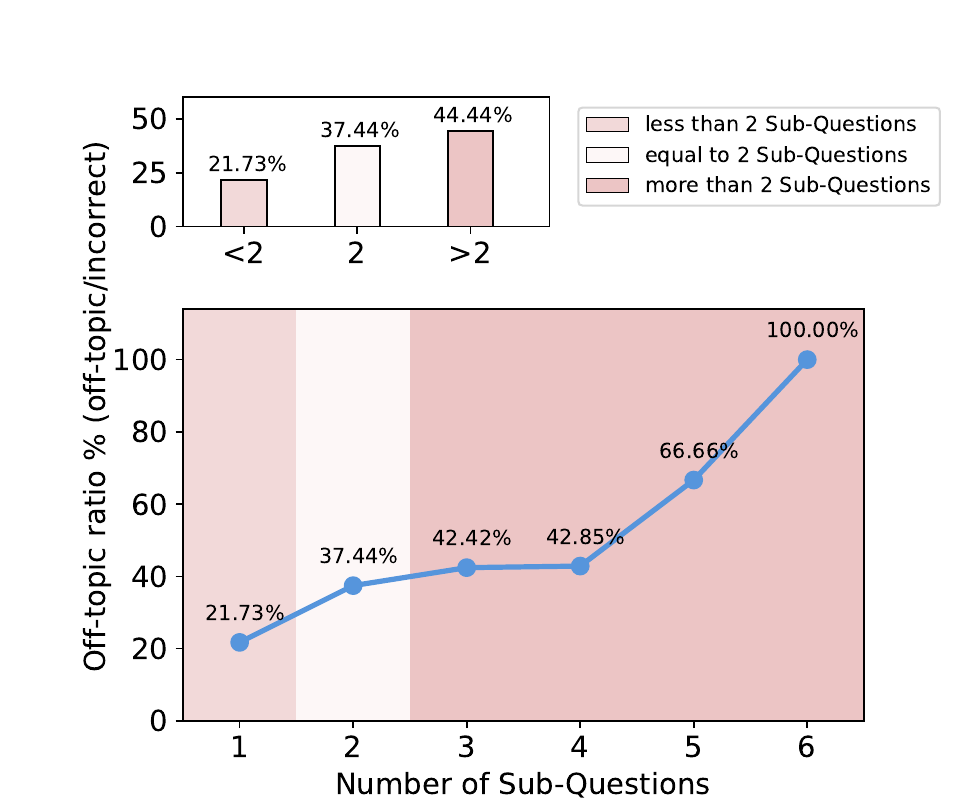}
\vspace*{-1em}
\caption{Off-topic ratio versus number of \texttt{Sub-Questions}. Evaluation conducted on 108 off-topic answers from the HotpotQA dataset.}
\label{fig:task_vs_mismatch}
\vspace*{-0.5em}
\end{figure}


In this subsection, we discuss the relationship between off-topic answers and the number of tasks in \texttt{Sub-Questions}.
As the line chart in Figure \ref{fig:task_vs_mismatch} illustrates, the off-topic ratio increases monotonically as the number of tasks rises.
Ideally, the reasoning chain should contain exactly 2 \texttt{Sub-Questions}, since the questions from the HotpotQA dataset are designed to be 2-hop.
Accordingly, depending on whether the number of tasks is less than, equal to, or greater than 2, we organize the same data into three groups in the bar chart of Figure \ref{fig:task_vs_mismatch}.
For the 1-hop reasoning chains, in which off-topic answers account for 21.73\% of cases, we notice that ReAct+ tends to conclude the answer prematurely if it is on-topic.
For reasoning chains with more than 2 hops, where off-topic answers occur 44.44\% of the time, we observe that ReAct+ becomes trapped in self-correction cycles.
This unavoidably adds noise to the context, causing the original question to be forgotten and resulting in off-topic answers.

\subsection{The Effect of Question Types}

\begin{figure}[htb]
\centering
\vspace*{-2em}
\includegraphics[width=\linewidth]{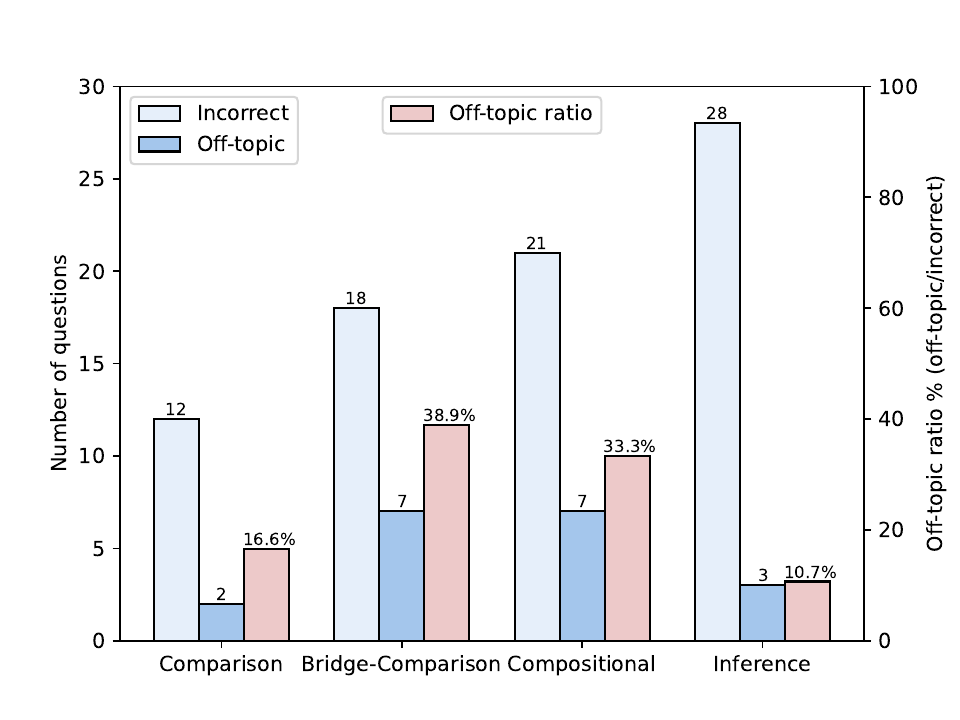}
\vspace*{-1.5em}
\caption{Off-topic answers versus question types. We randomly sample 30 questions per type on the 2WikiMultiHopQA dataset.}
\label{fig:question_type}
\end{figure}

In this subsection, we examine the relationship between off-topic answers and the four question types on the 2WikiMultiHopQA dataset.
The results are shown in Figure \ref{fig:question_type}.
The Comparison type has the second lowest off-topic ratio, as the answer is already provided within the original question.
For example, in a question like ``Who is older, A or B?'', the answer can only be either A or B, both options being clearly stated in the question itself.
The Bridge-Comparison type has the highest off-topic ratio.
For instance, in a question like ``Which film has the director born first, A or B?'', ReAct+ is prone to answering the name of the director rather than the name of the film, even though the correct answer must be either A or B.
We hypothesize that the Bridge-Comparison type requires a minimum of 4 tasks to be properly solved, as extended reasoning causes the original question to be forgotten.
The Inference type exhibits the lowest off-topic ratio but the highest error ratio.
We observe that ReAct+ tends to prematurely terminate the reasoning chain when an intermediate answer is on-topic. 
For example, in a question like ``Who is A's grandfather?'', where the first retrieved passage contains A's father B, the corresponding conclusion might erroneously regard B as A's grandfather.

%% file: 5_related_work.tex
\section{Related Work}

\textbf{Large Language Models on Open Domain Multi-Hop Question Answering.}
LLMs have shown outstanding capabilities in reasoning, planning, and utilizing tools, all of which are indispensable for performing natural language question answering \citep{he2022rethinking,trivedi2022interleaving}.
The pioneering work CoT \citep{wei2022chain} proposed to generate intermediate reasoning steps to answer questions that require complex reasoning capabilities.
Inspired by CoT, Self-Consistency \citep{wang2023self} and Complexity-Based Prompting \citep{fu2022complexity} sampled multiple reasoning trajectories to vote for the final answer.
Despite their performance enhancement, these voting-based CoT methods are resource-intensive and time-consuming.
Meanwhile, \citet{press2022measuring} pointed out that answering a compositional question as a whole is often more difficult than correctly answering its individual sub-questions.
Based on this insight, Self-Ask \citep{press2022measuring}, PS \citep{wang2023plan}, ReAct \citep{yao2022react}, and DSP \citep{khattab2022demonstrate} decomposed the compositional question into a series of sub-questions and employed external knowledge to prevent factual errors when answering individual sub-questions.
Our Dr3 is constructed based on the ReAct framework, as it was the most influential among the aforementioned works and also inspired subsequent LLM-Agent research \citep{mialon2023augmented,wang2023survey}.

\textbf{Post-Hoc Correction.}
Post-hoc correction refers to the process of refining the output from LLMs after it has been generated, without making any modification to the model parameters \citep{pan2023automatically}.
Self-Refine \citep{madaan2023self} iteratively polished outputs by incorporating LLM feedback for dialogue and code generation tasks.
Auto-Post-Editing \citep{raunak2023leveraging} and RCI \citep{kim2023language} demonstrated similar ideas of using LLMs to provide advice then using that advice to prompt higher quality outputs, which were applied to translation and computer tasks.
DIN-SQL \citep{pourreza2023din} proposed generic and gentle correction modules to fix bugs and potential issues respectively in the text-to-SQL task.
When it comes to post-hoc correction methods for QA tasks, Verify-and-Edit \citep{zhao2023verify} and LLM-AUGMENTER \citep{peng2023check} employed external knowledge to enhance the quality of the generated text.
In open-domain QA, SearChain \citep{xu2023search} utilized a trained small model to improve the quality of the retrieved passage.

While previous research concentrated on correcting factual errors in the reasoning process, our Dr3 focuses on alleviate the off-topic issue.
Furthermore, our Dr3 does not rely on any voting-based mechanism or additional tools for fact-checking sub-questions.

%% file: 6_conclusion.tex
\section{Conclusion}

In this paper, we have identified the crucial issue of off-topic answers that occurs when utilizing Large Language Models (LLMs) to tackle open domain multi-hop question answering (ODMHQA).
Our proposed solution, the Discriminate$\rightarrow$Re-Compose$\rightarrow$Re-Solve$\rightarrow$Re-Decompose (Dr3) mechanism, effectively harnesses the intrinsic capabilities of LLMs to detect and correct off-topic answers by performing step-wise revisions along the reversed reasoning chain.
Through comprehensive experiments conducted on the HotpotQA and 2WikiMultiHopQA datasets, we have demonstrated the effectiveness of our approach in alleviating the off-topic issue.
We anticipate that our work not only provides a practical solution to the issue of off-topic answers but also serves as a catalyst for future research in this area.

\section{Acknowledgments}
We would like to thank the anonymous reviewers for their valuable feedback. This work was supported by the NSFC (No. 62376120, 61936012, 62206126).

%% file: 7_appendix.tex
\clearpage
 

\section{ReAct+ Prompt}
\label{sec:react_plus_prompt}

\begin{table}[hb]
\renewcommand{\arraystretch}{1.4}
\centering  
\begin{tabularx}{\textwidth}{X} 
\toprule
\textcolor{gray}{\# Instruction} \\
You are a question-answering agent. To answer a difficult Question, you need to perform \textbf{Decomposition} to divide it into several tasks, solve them and integrate the information for the answer. \\
To solve each task, you need to use interleaving \textbf{Task}, \textbf{Action}, \textbf{Observation}, and \textbf{Conclusion} steps. The steps are:\\
    \quad 1. \textbf{Task}: a sub-problem to be solved from Decomposition and the previous Conclusion.\\
    \quad 2. \textbf{Action}: Search[Query] to retrieve a document corresponding to the Query.\\
    \quad 3. \textbf{Observation}: the retrieved document by the Action.\\
    \quad 4. \textbf{Conclusion}: the Task result according to the Observation.\\
According to the \textbf{Decomposition}, when all the necessary tasks are finished,
you need to execute \textbf{Composition} and then answer the Question with \textbf{Finish[Answer]}. The steps are: \\
    \quad 1. \textbf{Composition}: the composition of the information from all the tasks. \\
    \quad 2. \textbf{Finish[Answer]}: the final Answer to the Question.
\\
\\
\textcolor{gray}{\# We demonstrate a case here; all cases can be found in our codes.} \\
\textbf{Question}: Musician and satirist Allie Goertz wrote a song about the ``The Simpsons'' 
character Milhouse, who Matt Groening named after who?\\
\textbf{Decomposition}: The question simplifies to ``The Simpsons'' character Milhouse is named after who. I only need to search Milhouse and find who it is named after.\\
\textcolor{gray}{\# Sub-Question 1} \\
\textbf{Task 1}: I need to search Milhouse and find who it is named after.\\
\textbf{Action 1}: Search[Milhouse]\\
\textbf{Observation 1}: Milhouse Mussolini Van Houten is a recurring character in the Fox animated television series The Simpsons voiced by Pamela Hayden and created by Matt Groening.\\
\textbf{Conclusion 1}: The paragraph does not tell who Milhouse is named after.\\
\textcolor{gray}{\# Sub-Question 2} \\
\textbf{Task 2}: I can search Milhouse named after whom instead to find who it is named after.\\
\textbf{Action 2}: Search[Milhouse named after whom]\\
\textbf{Observation 2}: Milhouse was named after U.S. president Richard Nixon, whose middle name was 
Milhous.\\
\textbf{Conclusion 2}: Milhouse was named after U.S. president Richard Nixon.\\
\textbf{Composition}: Milhouse was named after U.S. president Richard Nixon, so the answer is Richard Nixon.\\
\textbf{Finish}: [Richard Nixon]
\\
\bottomrule
\end{tabularx}
\label{tab:react+_prompt} 
\end{table}

\clearpage

\section{Discriminator Prompt}
\label{sec:discriminator_prompt}

\begin{table}[h]
\renewcommand{\arraystretch}{1.4}
\centering  
\begin{tabularx}{\textwidth}{X}  
\toprule
\textcolor{gray}{\# Instruction} \\
You will be given QUESTION and ANSWER. You need to identify the possible answer to the QUESTION, and then check whether the ANSWER is the possible ANSWER.\\
For every case, you must think firstly in \textbf{THOUGHT}, and then output your \textbf{JUDGMENT}, with the format:\\
\textbf{THOUGHT}: (your analysis here).\\
\textbf{JUDGMENT}: YES / NO\\
\\

\textcolor{gray}{\# We demonstrate a case here; all cases can be found in our codes.} \\
\textbf{QUESTION}: When was the Man Falls in Love born on?\\
\textbf{ANSWER}: July 5, 1984\\
\textbf{THOUGHT}: The answer to the QUESTION can be a year, date or time range. The ANSWER ``July 5, 1984'' is a date. So the answer is YES.\\
\textbf{JUDGMENT}: YES\\
\bottomrule
\end{tabularx}
\label{tab:discriminator_prompt} 
\end{table}

\clearpage

\section{Re-Decompose Prompt}
\label{sec:re-decompose_prompt}

\begin{table}[h]
\renewcommand{\arraystretch}{1.4}
\centering  
\begin{tabularx}{\textwidth}{X}  
\toprule
\textcolor{gray}{\# Instruction} \\
You are a teacher, and your student is in the process of decomposing a multi-hop QUESTION into several sub-questions, solving them to arrive at the final answer.\\
However, there are some mistakes in the \textbf{STUDENT DECOMPOSITION} that the student cannot
answer the question. e.g., \textbf{misunderstanding the question}, \textbf{the divided ones are too ambiguous or difficult to lead to the final answer}, and so on.\\
For every case, you must analyze the question first in the \textbf{ANALYSIS}, which contains the understanding of the question, and the weaknesses of the student decomposition. After analysis, give a targeted suggestion based on the understanding of the problem and students' weaknesses in the \textbf{ADVICE}. Finally, refer to the advice, and give a more specific, correct and fluent decomposition in the \textbf{NEW DECOMPOSITION.}\\

The Input would be in the following format:\\
\textbf{QUESTION}: (the difficult question)\\
\textbf{STUDENT DECOMPOSITION}: (the decomposition from the student)\\
\\

The Output must be in the following format:\\
\textbf{ANALYSIS}: (analysis here. Analyze the question and student's decomposition)\\
\textbf{ADVICE}: (advice here. Refer to the analysis, give targeted advice)\\
\textbf{NEW DECOMPOSITION}: (new decomposition here. Refer to the advice, give a more specific, correct and fluent version than the STUDENT DECOMPOSITION)\\
\\

\textcolor{gray}{\# We demonstrate a case here; all cases can be found in our codes.} \\
\textbf{QUESTION}: Is Level 3 Communications headquartered further west than Advanced Micro Devices?\\
\textbf{STUDENT DECOMPOSITION}: I need to find the headquarters location of Advanced Micro Devices and Level 3 Communications, and then integrate the information to identify which company is headquartered further west for the final answer\\
\textbf{ANALYSIS}: The answer to the question should be either yes or no, if Level 3 Communications is headquartered further west than Advanced Micro Devices, the answer is yes, other is no. The student decomposition tries to "identify which company", the answer delivered from the decomposition will be a company, neither yes nor no, which causes the error.\\
\textbf{ADVICE}: I think the last step should find whether Level 3 Communications is headquartered further west, instead of "identify which company...".\\
\textbf{NEW DECOMPOSITION}:  I need to find the headquarters location of Advanced Micro Devices and Level 3 Communications, and then find whether Level 3 Communications is headquartered further west for the final answer.
\\
\bottomrule
\end{tabularx}
\label{tab:re_decompose_prompt} 
\end{table}

\clearpage

\section{Case Study}
\label{sec:special_case}

\begin{samepage}
\begin{table*}[b]
\renewcommand{\arraystretch}{1.15}
\centering  
\begin{tabularx}{\textwidth}{l|XX} 
\toprule
\textbf{Step} & \textbf{ReAct+} & \textbf{Dr3} \\
\hline
\textbf{Question} & \multicolumn{2}{>{\hsize=\dimexpr2\hsize+2\tabcolsep+1\arrayrulewidth\relax}X}{What nationality is the man who developed the dictator game? 
} \\ 
\\
\textbf{Decomposition} & \multicolumn{2}{>{\hsize=\dimexpr2\hsize+2\tabcolsep+1\arrayrulewidth\relax}X}{I need to find the nationality of the man who developed the dictator game, which is the final answer.
} \\
\\
\textbf{Sub-Question 1} & \multicolumn{2}{>{\hsize=\dimexpr2\hsize+2\tabcolsep+1\arrayrulewidth\relax}X}{- Find the man who developed the dictator game.\newline
- Answer: \textcolor{deepgreen}{Michael Pagano} \textcolor{gray}{\# Correct} 
} \\
\\
\textbf{Sub-Question 2} &\multicolumn{2}{>{\hsize=\dimexpr2\hsize+2\tabcolsep+1\arrayrulewidth\relax}X}{- Find the nationality of Michael Pagano.\newline
- Answer: \textcolor{red}{From Los Angeles, California} \textcolor{gray}{\# ``Los Angeles, California'' is a place, not a nationality.}
}\\
\cdashline{2-3}
\\
\textbf{Composition}& The man who developed the dictator game is Michael Pagano, Michael Pagano is from Los Angeles, California, \textcolor{red}{so the answer is Los Angeles, California.}
& \textcolor{purple}{No, the answer is not ``Los Angeles, California'', and} the answer should be the nationality, \textcolor{deepgreen}{so the answer is American.}\\ 
\\
\textbf{Answer}& \textcolor{red}{Los Angeles, California} & \textcolor{deepgreen}{American} \\

\hline
\textbf{Question} & \multicolumn{2}{>{\hsize=\dimexpr2\hsize+2\tabcolsep+1\arrayrulewidth\relax}X}{Philip Savage served as Director of Player Personnel for the Baltimore Ravens under what general manager who was inducted into both the College and Pro Football Halls of Fame?} \\ 
\\
\textbf{Decomposition} & \multicolumn{2}{>{\hsize=\dimexpr2\hsize+2\tabcolsep+1\arrayrulewidth\relax}X}{I need to find the general manager of Baltimore Ravens during Philip Savage served as Director of Player Personnel, and then \textcolor{red}{find whether the general manager was inducted into both the College and Pro Football Halls of Fame for the final answer.} \textcolor{gray}{\# The answer should be the person who satisfies the two requirements.}
} \\
\\
\textbf{Sub-Question 1} & \multicolumn{2}{>{\hsize=\dimexpr2\hsize+2\tabcolsep+1\arrayrulewidth\relax}X}{- Find the general manager of Baltimore Ravens during he served as Director of Player Personnel.\newline
- Answer: \textcolor{deepgreen}{Ozzie Newsome} \textcolor{gray}{\# Correct} 
} \\
\\
\textbf{Sub-Question 2} &\multicolumn{2}{>{\hsize=\dimexpr2\hsize+2\tabcolsep+1\arrayrulewidth\relax}X}{- Find whether he was inducted into both the College and Pro Football Halls of Fame\newline
- Answer: \textcolor{deepgreen}{Yes} \textcolor{gray}{\# Correct}
}\\
\cdashline{2-3}
\\
\textbf{Composition}& The general manager is Ozzie Newsome, who was inducted into both the College and Pro Football Halls of Fame, \textcolor{red}{so the answer is Yes.}
& \textcolor{purple}{No, the answer is not ``Yes'', and} Ozzie Newsome is the manager and was inducted into both the College and Pro
Football Halls of Fame, \textcolor{deepgreen}{so the answer is Ozzie Newsome.}\\ 
\\
\textbf{Answer}& \textcolor{red}{Yes} & \textcolor{deepgreen}{Ozzie Newsome} \\
\bottomrule
\end{tabularx}
\caption{Two cases where the error is fixed in the Re-Compose stage. The error in the upper case occurs in the \texttt{Sub-Question} stage, while the error in the lower case occurs in the \texttt{Decomposition} stage.}
\label{tab:special_cases}
\end{table*}

In Table \ref{tab:special_cases}, we demonstrate two cases in which off-topic answers are corrected by our Dr3 mechanism.
Specifically, these two cases differ in that the error in the upper case occurs in the \texttt{Sub-Question} stage, while the error in the lower case occurs in the \texttt{Decomposition} stage.
However, both can be corrected by the same Re-Compose strategy of our Dr3 mechanism.
\end{samepage}

%% file: lrec-coling2024-example.bbl
\begin{thebibliography}{40}
\expandafter\ifx\csname natexlab\endcsname\relax\def\natexlab#1{#1}\fi

\bibitem[{Bang et~al.(2023)Bang, Cahyawijaya, Lee, Dai, Su, Wilie, Lovenia, Ji,
  Yu, Chung et~al.}]{bang2023multitask}
Yejin Bang, Samuel Cahyawijaya, Nayeon Lee, Wenliang Dai, Dan Su, Bryan Wilie,
  Holy Lovenia, Ziwei Ji, Tiezheng Yu, Willy Chung, et~al. 2023.
\newblock A multitask, multilingual, multimodal evaluation of chatgpt on
  reasoning, hallucination, and interactivity.
\newblock \emph{arXiv preprint arXiv:2302.04023}.

\bibitem[{Brown et~al.(2020)Brown, Mann, Ryder, Subbiah, Kaplan, Dhariwal,
  Neelakantan, Shyam, Sastry, Askell et~al.}]{brown2020language}
Tom Brown, Benjamin Mann, Nick Ryder, Melanie Subbiah, Jared~D Kaplan, Prafulla
  Dhariwal, Arvind Neelakantan, Pranav Shyam, Girish Sastry, Amanda Askell,
  et~al. 2020.
\newblock Language models are few-shot learners.
\newblock \emph{Advances in Neural Information Processing Systems},
  33:1877--1901.

\bibitem[{Feldman and El-Yaniv(2019)}]{feldman2019multi}
Yair Feldman and Ran El-Yaniv. 2019.
\newblock Multi-hop paragraph retrieval for open-domain question answering.
\newblock In \emph{Proceedings of the 57th Annual Meeting of the Association
  for Computational Linguistics}, pages 2296--2309.

\bibitem[{Fu et~al.(2022)Fu, Peng, Sabharwal, Clark, and
  Khot}]{fu2022complexity}
Yao Fu, Hao Peng, Ashish Sabharwal, Peter Clark, and Tushar Khot. 2022.
\newblock Complexity-based prompting for multi-step reasoning.
\newblock In \emph{The 11th International Conference on Learning
  Representations}.

\bibitem[{Hao et~al.(2023)Hao, Liu, Wang, and Hu}]{hao2023toolkengpt}
Shibo Hao, Tianyang Liu, Zhen Wang, and Zhiting Hu. 2023.
\newblock Toolkengpt: Augmenting frozen language models with massive tools via
  tool embeddings.
\newblock \emph{arXiv preprint arXiv:2305.11554}.

\bibitem[{He et~al.(2022)He, Zhang, and Roth}]{he2022rethinking}
Hangfeng He, Hongming Zhang, and Dan Roth. 2022.
\newblock Rethinking with retrieval: Faithful large language model inference.
\newblock \emph{arXiv preprint arXiv:2301.00303}.

\bibitem[{Ho et~al.(2020)Ho, Nguyen, Sugawara, and Aizawa}]{ho2020constructing}
Xanh Ho, Anh-Khoa~Duong Nguyen, Saku Sugawara, and Akiko Aizawa. 2020.
\newblock Constructing a multi-hop qa dataset for comprehensive evaluation of
  reasoning steps.
\newblock In \emph{Proceedings of the 28th International Conference on
  Computational Linguistics}, pages 6609--6625.

\bibitem[{Hsieh et~al.(2023)Hsieh, Chen, Li, Fujii, Ratner, Lee, Krishna, and
  Pfister}]{hsieh2023tool}
Cheng-Yu Hsieh, Si-An Chen, Chun-Liang Li, Yasuhisa Fujii, Alexander Ratner,
  Chen-Yu Lee, Ranjay Krishna, and Tomas Pfister. 2023.
\newblock Tool documentation enables zero-shot tool-usage with large language
  models.
\newblock \emph{arXiv preprint arXiv:2308.00675}.

\bibitem[{Khattab et~al.(2022)Khattab, Santhanam, Li, Hall, Liang, Potts, and
  Zaharia}]{khattab2022demonstrate}
Omar Khattab, Keshav Santhanam, Xiang~Lisa Li, David Hall, Percy Liang,
  Christopher Potts, and Matei Zaharia. 2022.
\newblock Demonstrate-search-predict: Composing retrieval and language models
  for knowledge-intensive nlp.
\newblock \emph{arXiv preprint arXiv:2212.14024}.

\bibitem[{Kim et~al.(2023)Kim, Baldi, and McAleer}]{kim2023language}
Geunwoo Kim, Pierre Baldi, and Stephen McAleer. 2023.
\newblock Language models can solve computer tasks.
\newblock \emph{arXiv preprint arXiv:2303.17491}.

\bibitem[{Kojima et~al.(2022)Kojima, Gu, Reid, Matsuo, and
  Iwasawa}]{kojima2022large}
Takeshi Kojima, Shixiang~Shane Gu, Machel Reid, Yutaka Matsuo, and Yusuke
  Iwasawa. 2022.
\newblock Large language models are zero-shot reasoners.
\newblock \emph{Advances in Neural Information Processing Systems},
  35:22199--22213.

\bibitem[{Louis and Higgins(2010)}]{louis2010off}
Annie Louis and Derrick Higgins. 2010.
\newblock Off-topic essay detection using short prompt texts.
\newblock In \emph{Proceedings of the NAACL HLT 2010 5th Workshop on Innovative
  Use of NLP for Building Educational Applications}, pages 92--95.

\bibitem[{Lyu et~al.(2023)Lyu, Havaldar, Stein, Zhang, Rao, Wong, Apidianaki,
  and Callison-Burch}]{lyu2023faithful}
Qing Lyu, Shreya Havaldar, Adam Stein, Li~Zhang, Delip Rao, Eric Wong, Marianna
  Apidianaki, and Chris Callison-Burch. 2023.
\newblock Faithful chain-of-thought reasoning.
\newblock \emph{arXiv preprint arXiv:2301.13379}.

\bibitem[{Madaan et~al.(2023)Madaan, Tandon, Gupta, Hallinan, Gao, Wiegreffe,
  Alon, Dziri, Prabhumoye, Yang et~al.}]{madaan2023self}
Aman Madaan, Niket Tandon, Prakhar Gupta, Skyler Hallinan, Luyu Gao, Sarah
  Wiegreffe, Uri Alon, Nouha Dziri, Shrimai Prabhumoye, Yiming Yang, et~al.
  2023.
\newblock Self-refine: Iterative refinement with self-feedback.
\newblock \emph{arXiv preprint arXiv:2303.17651}.

\bibitem[{Malinin et~al.(2016)Malinin, Van~Dalen, Knill, Wang, and
  Gales}]{malinin2016off}
Andrey Malinin, Rogier Van~Dalen, Kate Knill, Yu~Wang, and Mark Gales. 2016.
\newblock Off-topic response detection for spontaneous spoken english
  assessment.
\newblock In \emph{Proceedings of the 54th Annual Meeting of the Association
  for Computational Linguistics}, volume~1, pages 1075--1084.

\bibitem[{Mavi et~al.(2022)Mavi, Jangra, and Jatowt}]{mavi2022survey}
Vaibhav Mavi, Anubhav Jangra, and Adam Jatowt. 2022.
\newblock A survey on multi-hop question answering and generation.
\newblock \emph{arXiv preprint arXiv:2204.09140}.

\bibitem[{Mialon et~al.(2023)Mialon, Dess{\`\i}, Lomeli, Nalmpantis, Pasunuru,
  Raileanu, Rozi{\`e}re, Schick, Dwivedi-Yu, Celikyilmaz
  et~al.}]{mialon2023augmented}
Gr{\'e}goire Mialon, Roberto Dess{\`\i}, Maria Lomeli, Christoforos Nalmpantis,
  Ram Pasunuru, Roberta Raileanu, Baptiste Rozi{\`e}re, Timo Schick, Jane
  Dwivedi-Yu, Asli Celikyilmaz, et~al. 2023.
\newblock Augmented language models: a survey.
\newblock \emph{arXiv preprint arXiv:2302.07842}.

\bibitem[{Mishra et~al.(2022)Mishra, Khashabi, Baral, Choi, and
  Hajishirzi}]{mishra2022reframing}
Swaroop Mishra, Daniel Khashabi, Chitta Baral, Yejin Choi, and Hannaneh
  Hajishirzi. 2022.
\newblock Reframing instructional prompts to gptk's language.
\newblock In \emph{Proceedings of the 60th Annual Meeting of the Association
  for Computational Linguistics}, pages 589--612.

\bibitem[{Nakano et~al.(2021)Nakano, Hilton, Balaji, Wu, Ouyang, Kim, Hesse,
  Jain, Kosaraju, Saunders et~al.}]{nakano2021webgpt}
Reiichiro Nakano, Jacob Hilton, Suchir Balaji, Jeff Wu, Long Ouyang, Christina
  Kim, Christopher Hesse, Shantanu Jain, Vineet Kosaraju, William Saunders,
  et~al. 2021.
\newblock Webgpt: Browser-assisted question-answering with human feedback.
\newblock \emph{arXiv preprint arXiv:2112.09332}.

\bibitem[{Pan et~al.(2023)Pan, Saxon, Xu, Nathani, Wang, and
  Wang}]{pan2023automatically}
Liangming Pan, Michael Saxon, Wenda Xu, Deepak Nathani, Xinyi Wang, and
  William~Yang Wang. 2023.
\newblock Automatically correcting large language models: Surveying the
  landscape of diverse self-correction strategies.
\newblock \emph{arXiv preprint arXiv:2308.03188}.

\bibitem[{Peng et~al.(2023)Peng, Galley, He, Cheng, Xie, Hu, Huang, Liden, Yu,
  Chen et~al.}]{peng2023check}
Baolin Peng, Michel Galley, Pengcheng He, Hao Cheng, Yujia Xie, Yu~Hu, Qiuyuan
  Huang, Lars Liden, Zhou Yu, Weizhu Chen, et~al. 2023.
\newblock Check your facts and try again: Improving large language models with
  external knowledge and automated feedback.
\newblock \emph{arXiv preprint arXiv:2302.12813}.

\bibitem[{Pourreza and Rafiei(2023)}]{pourreza2023din}
Mohammadreza Pourreza and Davood Rafiei. 2023.
\newblock Din-sql: Decomposed in-context learning of text-to-sql with
  self-correction.
\newblock \emph{arXiv preprint arXiv:2304.11015}.

\bibitem[{Press et~al.(2022)Press, Zhang, Min, Schmidt, Smith, and
  Lewis}]{press2022measuring}
Ofir Press, Muru Zhang, Sewon Min, Ludwig Schmidt, Noah~A Smith, and Mike
  Lewis. 2022.
\newblock Measuring and narrowing the compositionality gap in language models.
\newblock \emph{arXiv preprint arXiv:2210.03350}.

\bibitem[{Rajpurkar et~al.(2016)Rajpurkar, Zhang, Lopyrev, and
  Liang}]{rajpurkar2016squad}
Pranav Rajpurkar, Jian Zhang, Konstantin Lopyrev, and Percy Liang. 2016.
\newblock Squad: 100,000+ questions for machine comprehension of text.
\newblock In \emph{Proceedings of the 2016 Conference on Empirical Methods in
  Natural Language Processing}, pages 2383--2392.

\bibitem[{Raunak et~al.(2023)Raunak, Sharaf, Awadallah, and
  Menezes}]{raunak2023leveraging}
Vikas Raunak, Amr Sharaf, Hany~Hassan Awadallah, and Arul Menezes. 2023.
\newblock Leveraging gpt-4 for automatic translation post-editing.
\newblock \emph{arXiv preprint arXiv:2305.14878}.

\bibitem[{Ruan et~al.(2023)Ruan, Chen, Zhang, Xu, Bao, Du, Shi, Mao, Zeng, and
  Zhao}]{ruan2023tptu}
Jingqing Ruan, Yihong Chen, Bin Zhang, Zhiwei Xu, Tianpeng Bao, Guoqing Du,
  Shiwei Shi, Hangyu Mao, Xingyu Zeng, and Rui Zhao. 2023.
\newblock Tptu: Task planning and tool usage of large language model-based ai
  agents.
\newblock \emph{arXiv preprint arXiv:2308.03427}.

\bibitem[{Santhanam et~al.(2022)Santhanam, Khattab, Saad-Falcon, Potts, and
  Zaharia}]{santhanam2022colbertv2}
Keshav Santhanam, Omar Khattab, Jon Saad-Falcon, Christopher Potts, and Matei
  Zaharia. 2022.
\newblock Colbertv2: Effective and efficient retrieval via lightweight late
  interaction.
\newblock In \emph{Proceedings of the 2022 Conference of the North American
  Chapter of the Association for Computational Linguistics: Human Language
  Technologies}, pages 3715--3734.

\bibitem[{Trivedi et~al.(2022)Trivedi, Balasubramanian, Khot, and
  Sabharwal}]{trivedi2022interleaving}
Harsh Trivedi, Niranjan Balasubramanian, Tushar Khot, and Ashish Sabharwal.
  2022.
\newblock Interleaving retrieval with chain-of-thought reasoning for
  knowledge-intensive multi-step questions.
\newblock \emph{Proceedings of the 60th Annual Meeting of the Association for
  Computational Linguistics}.

\bibitem[{Wang et~al.(2023{\natexlab{a}})Wang, Ma, Feng, Zhang, Yang, Zhang,
  Chen, Tang, Chen, Lin et~al.}]{wang2023survey}
Lei Wang, Chen Ma, Xueyang Feng, Zeyu Zhang, Hao Yang, Jingsen Zhang, Zhiyuan
  Chen, Jiakai Tang, Xu~Chen, Yankai Lin, et~al. 2023{\natexlab{a}}.
\newblock A survey on large language model based autonomous agents.
\newblock \emph{arXiv preprint arXiv:2308.11432}.

\bibitem[{Wang et~al.(2023{\natexlab{b}})Wang, Xu, Lan, Hu, Lan, Lee, and
  Lim}]{wang2023plan}
Lei Wang, Wanyu Xu, Yihuai Lan, Zhiqiang Hu, Yunshi Lan, Roy Ka-Wei Lee, and
  Ee-Peng Lim. 2023{\natexlab{b}}.
\newblock Plan-and-solve prompting: Improving zero-shot chain-of-thought
  reasoning by large language models.
\newblock In \emph{Proceedings of the 61st Annual Meeting of the Association
  for Computational Linguistics}.

\bibitem[{Wang et~al.(2023{\natexlab{c}})Wang, Wei, Schuurmans, Le, Chi,
  Narang, Chowdhery, and Zhou}]{wang2023self}
Xuezhi Wang, Jason Wei, Dale Schuurmans, Quoc Le, Ed~Chi, Sharan Narang,
  Aakanksha Chowdhery, and Denny Zhou. 2023{\natexlab{c}}.
\newblock Self-consistency improves chain of thought reasoning in language
  models.
\newblock In \emph{The 11th International Conference on Learning
  Representations}.

\bibitem[{Wei et~al.(2022)Wei, Wang, Schuurmans, Bosma, Xia, Chi, Le, Zhou
  et~al.}]{wei2022chain}
Jason Wei, Xuezhi Wang, Dale Schuurmans, Maarten Bosma, Fei Xia, Ed~Chi, Quoc~V
  Le, Denny Zhou, et~al. 2022.
\newblock Chain-of-thought prompting elicits reasoning in large language
  models.
\newblock \emph{Advances in Neural Information Processing Systems},
  35:24824--24837.

\bibitem[{Xi et~al.(2023)Xi, Jin, Zhou, Zheng, Gao, Gui, Zhang, and
  Huang}]{xi2023self}
Zhiheng Xi, Senjie Jin, Yuhao Zhou, Rui Zheng, Songyang Gao, Tao Gui, Qi~Zhang,
  and Xuanjing Huang. 2023.
\newblock Self-polish: Enhance reasoning in large language models via problem
  refinement.
\newblock \emph{arXiv preprint arXiv:2305.14497}.

\bibitem[{Xu et~al.(2023)Xu, Pang, Shen, Cheng, and Chua}]{xu2023search}
Shicheng Xu, Liang Pang, Huawei Shen, Xueqi Cheng, and Tat-seng Chua. 2023.
\newblock Search-in-the-chain: Towards the accurate, credible and traceable
  content generation for complex knowledge-intensive tasks.
\newblock \emph{arXiv preprint arXiv:2304.14732}.

\bibitem[{Yang et~al.(2018)Yang, Qi, Zhang, Bengio, Cohen, Salakhutdinov, and
  Manning}]{yang2018hotpotqa}
Zhilin Yang, Peng Qi, Saizheng Zhang, Yoshua Bengio, William Cohen, Ruslan
  Salakhutdinov, and Christopher~D Manning. 2018.
\newblock Hotpotqa: A dataset for diverse, explainable multi-hop question
  answering.
\newblock In \emph{Proceedings of the 2018 Conference on Empirical Methods in
  Natural Language Processing}, pages 2369--2380.

\bibitem[{Yao et~al.(2022)Yao, Zhao, Yu, Du, Shafran, Narasimhan, and
  Cao}]{yao2022react}
Shunyu Yao, Jeffrey Zhao, Dian Yu, Nan Du, Izhak Shafran, Karthik~R Narasimhan,
  and Yuan Cao. 2022.
\newblock React: Synergizing reasoning and acting in language models.
\newblock In \emph{The 11th International Conference on Learning
  Representations}.

\bibitem[{Zhang et~al.(2023)Zhang, Li, Cui, Cai, Liu, Fu, Huang, Zhao, Zhang,
  Chen, Wang, Luu, Bi, Shi, and Shi}]{zhang2023hallucination}
Yue Zhang, Yafu Li, Leyang Cui, Deng Cai, Lemao Liu, Tingchen Fu, Xinting
  Huang, Enbo Zhao, Yu~Zhang, Yulong Chen, Longyue Wang, Anh~Tuan Luu, Wei Bi,
  Freda Shi, and Shuming Shi. 2023.
\newblock Siren's song in the ai ocean: A survey on hallucination in large
  language models.
\newblock \emph{arXiv preprint arXiv:2309.01219}.

\bibitem[{Zhao et~al.(2023)Zhao, Li, Joty, Qin, and Bing}]{zhao2023verify}
Ruochen Zhao, Xingxuan Li, Shafiq Joty, Chengwei Qin, and Lidong Bing. 2023.
\newblock Verify-and-edit: A knowledge-enhanced chain-of-thought framework.
\newblock In \emph{Proceedings of the 61st Annual Meeting of the Association
  for Computational Linguistics}, volume~1.

\bibitem[{Zheng et~al.(2023)Zheng, Liu, Xie, Li, and Li}]{zheng2023progressive}
Chuanyang Zheng, Zhengying Liu, Enze Xie, Zhenguo Li, and Yu~Li. 2023.
\newblock Progressive-hint prompting improves reasoning in large language
  models.
\newblock \emph{arXiv preprint arXiv:2304.09797}.

\bibitem[{Zhu et~al.(2021)Zhu, Lei, Wang, Zheng, Poria, and
  Chua}]{zhu2021retrieving}
Fengbin Zhu, Wenqiang Lei, Chao Wang, Jianming Zheng, Soujanya Poria, and
  Tat-Seng Chua. 2021.
\newblock Retrieving and reading: A comprehensive survey on open-domain
  question answering.
\newblock \emph{arXiv preprint arXiv:2101.00774}.

\end{thebibliography}
